\documentclass[10pt,twocolumn,letterpaper]{article}

\usepackage{iccv}
\usepackage{times}
\usepackage{epsfig}
\usepackage{graphicx}
\usepackage{amsmath}
\usepackage{amssymb}

\usepackage{authblk}
\usepackage[toc,page]{appendix}

\usepackage[pagebackref=true,breaklinks=true,letterpaper=true,colorlinks,bookmarks=false]{hyperref}
\usepackage{multirow}
\iccvfinalcopy 

\def\iccvPaperID{***} 

\ificcvfinal\fi
\begin{document}

\title{Cross-domain Image Retrieval with a Dual Attribute-aware Ranking Network}

\author[1]{Junshi Huang}
\author[2]{Rogerio S. Feris}
\author[2]{Qiang Chen}
\author[1]{Shuicheng Yan}
\affil[1]{National University of Singapore, Singapore}
\affil[2]{IBM Research}
\affil[ ]{\tt\small {\{a0092558, eleyans\}@nus.edu.sg}}
\affil[ ]{\tt\small {rsferis@us.ibm.com}}
\affil[ ]{\tt\small {qiangchen@au1.ibm.com}}

\maketitle

\begin{abstract}

We address the problem of cross-domain image retrieval, considering the following practical application: given a user photo depicting a clothing image, our goal is to retrieve the same or attribute-similar clothing items from online shopping stores. This is a challenging problem due to the large  discrepancy between online shopping images, usually taken in ideal lighting/pose/background conditions, and user photos captured in uncontrolled conditions. To address this problem, we propose a Dual Attribute-aware Ranking Network (DARN) for retrieval feature learning. 
More specifically, DARN consists of two sub-networks, one for each domain, whose retrieval feature representations are driven by semantic attribute learning. We show that this attribute-guided learning is a key factor for retrieval accuracy improvement. In addition, to further align with the nature of the retrieval problem, we impose a triplet visual similarity constraint for learning to rank across the two sub-networks.
Another contribution of our work is a large-scale dataset which makes the network learning feasible. We exploit customer review websites to crawl
a large set of online shopping images and corresponding offline user photos with fine-grained clothing attributes, \ie, around 450,000 online shopping images and about 90,000 exact offline counterpart images of those online ones. All these images are collected from real-world consumer websites reflecting the diversity of the data modality, which makes this dataset unique and rare in the academic community. We extensively evaluate the retrieval performance of networks in different configurations. The top-20 retrieval accuracy is \textbf{doubled} when using  the proposed DARN  other than the current popular solution using pre-trained CNN features only (0.570 vs. 0.268).

\end{abstract}

\vspace{-0.2in}
\section{Introduction}

\begin{figure}
\begin{center}
\includegraphics[width=0.45\textwidth,keepaspectratio]{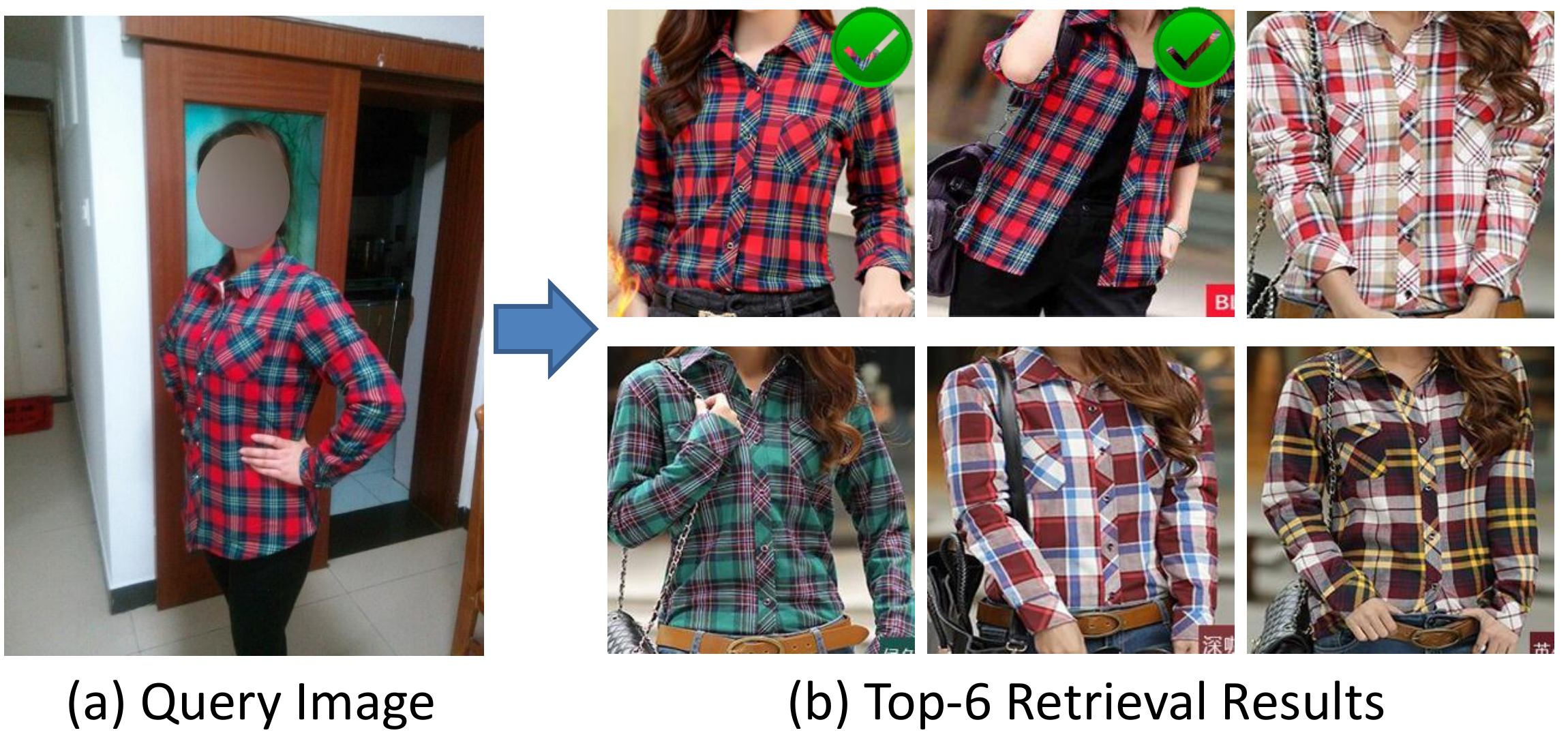}
\end{center}
\vspace{-0.1in}
\caption{Cross-domain clothing retrieval. (a) Query image from daily photos. (b) Top-6 product retrieval results from the online shopping domain. The proposed system finds the exact match clothing (first two images) and ranks the ones with similar attributes as top results.}
\label{fig:framework}
\vspace{-0.2in}

\end{figure}
\vspace{-0.1in}

There is a long history of methods for content-based image retrieval  in the field of computer vision \cite{QBIC, neuraleccv2014}. However, little work has been devoted to the problem of cross-domain image retrieval, defined as follows: given a query image from domain $X$, retrieve  similar images from a database of images belonging to domain $Y$. 

This problem setting arises in many important applications. For example, mobile product image search \cite{shen2012mobile} aims at identifying a product, or retrieving similar products from the online shopping domain based on a photo captured in unconstrained scenarios by a mobile phone camera. In surveillance applications, a security guard may be interested in retrieving images of a suspect from a specific camera given a query image from another camera.

In this paper, we address the problem of cross-domain product retrieval by taking clothing 
products as a concrete use case. Given an offline clothing image from the ``street'' domain, our goal is to retrieve the same or similar clothing items from a large-scale gallery of professional online shopping images, as illustrated in Figure \ref{fig:framework}.

Due to the huge impact for e-commerce applications, there is a growing interest in methods for clothing retrieval \cite{kalantidis2013,manfredi2013,liu2012street,wang2011clothes} and outfit recommendation \cite{ebay2014}.
The majority of these methods, however,  do not model the discrepancy between the user photos and clothing images from online shopping stores. Another barrier also occurs because of the lack of large annotated training sets containing user photos and desired retrieved images from online shopping. 

In order to tackle the training data issue, we observe that there is a large number of customer review websites, where people post their pictures wearing the clothing they have purchased. Therefore, it is possible to crawl the offline clothing images uploaded by the users with the links to the online shopping product images. As a result, we created a dataset containing tens of thousands of online-offline clothing image pairs obtained from the user review pages. These image pairs are very rare in both academic and industry as they reveal the real discrepancy of images across scenarios. In addition, we have also obtained corresponding fine-grained clothing attributes (\eg, clothing color, collar pattern, sleeve shape, sleeve length, \etc.) from the available online product description, without significant annotation cost.
As data pre-processing, in order to remove the impact of cluttered backgrounds, which predominantly exist for the offline images, we employ an enhanced R-CNN detector to localize the clothing area in the image,  with some refinements particularly made for the clothing detection problem.




For addressing the problem of cross-domain retrieval, we propose a novel Dual Attribute-aware Ranking Network (DARN) for retrieval feature learning.  DARN consists of two sub-networks  with similar  structure.  Each of the two domain images are fed into each of the two sub-networks. This specific design aims to diminish the discrepancy of online and offline images.


The two sub-networks are designed to be driven by semantic attribute learning, so we call them attribute-aware networks. The intuition is to create a powerful semantic representation of clothing in each domain, by leveraging the vast amounts of data annotated with fine-grained clothing attributes. Tree-structure layers are embedded into each sub-network for the comprehensive integration of attributes and their full relations. Specifically, the low-level layers of the sub-network are shared for learning the low-level representation. Then, a set of fully connected layers in a tree-structure are used to construct the high-level component, with each branch modelling one attribute.

Based on the learned semantic features from each attribute-aware network, we incorporate the learning-to-rank objective to further enhance the retrieval feature representation. Specifically, the triplet ranking loss is used to constrain the feature similarity of triplets, \ie,  the feature distance between the online-offline image pair must be smaller than that of offline image and any other dissimilar online images.

Generally, the retrieval features from DARN have several advantages compared with the deep features of other works \cite{jia2014caffe, donahue2013decaf}.
(1) By using the dual-structure network, our model can handle the cross-domain problem more appropriately. (2)
In each sub-network, the scenario-specific semantic representation of clothing is elaborately captured by leveraging the tree-structure layers.
(3) Based on the semantic representation, the visual similarity constraint enables more effective feature learning for the retrieval problem.

In summary, the main {\bf contributions} of our paper are:

\begin{enumerate}
\item We collect a unique dataset composed of cross-scenario image pairs with fine-grained attributes. The number of online images is about 450,000, with additional 90,000 offline counterparts collected. Each image has about 5-9 semantic attribute categories, with more than a hundred possible attribute values.
This online-offline image pair dataset provides a training/testing platform for many real-world applications related to clothing analytics.
We are planning to release the full dataset to the community for research purposes only.

\vspace{-0.05in}


\item We propose the Dual Attribute-Aware Ranking Network  which simultaneously integrates the attributes and visual similarity constraint into the retrieval feature learning. We design  tree-structure layers to comprehensively capture the  attribute information and their full relations, which provides a new insight on multi-label learning. We also introduce the triplet loss function which perfectly fits into the deep network training.

\vspace{-0.05in}

\item We conduct extensive experiments proving the effectiveness and robustness of the framework and each one of its components for the clothing retrieval problem. The top-20 retrieval accuracy is \textbf{doubled} when using  the proposed DARN  other than  using pre-trained CNN feature only (0.570 vs. 0.268).  The proposed method is general and could be applied to  other cross-domain  image retrieval problems.


\vspace{-0.05in}
 
\vspace{-0.05in}


\end{enumerate}

\section{Related Work}

{\bf Fashion Datasets}.
Recently, several datasets containing a wide variety of clothing images captured from fashion websites have been carefully annotated with  attribute labels \cite{yamaguchi2012parsing, dongparsing2014,fashion10000,ebay2014}.
These datasets are primarily designed for training and evaluation of clothing parsing and attribute estimation algorithms. 
In contrast, our data is comprised of a large set of clothing image pairs depicting user photos and corresponding garments from online shopping, in addition to fine-grained attributes. Notably, this real-world data is essential to bridge the gap between the two domains.

{\bf Visual Analysis of Clothing}.
Many methods have been recently proposed for automated analysis of clothing images, spanning a wide range of application domains.
In particular, clothing recognition has been used for context-aided people identification \cite{andrew2008}, fashion style recognition \cite{hipster2014}, occupation recognition \cite{occupation2011}, and social tribe prediction \cite{tribe2013}. Clothing parsing methods, which produce semantic labels for each pixel in the input image, have received significant attention in the past few years \cite{yamaguchi2012parsing,dongparsing2014}.
In the surveillance domain, matching clothing images across cameras is a fundamental task for the well-known person re-identification problem \cite{gong2012, shi2015}. 

Recently, there is a growing interest in methods for clothing retrieval \cite{kalantidis2013,manfredi2013,liu2012street,wang2011clothes} and outfit recommendation \cite{ebay2014}.
Most of those methods do not model the discrepancy between the user photos and online clothing images. An exception is the work of Liu et al \cite{liu2012street}, which follows a very different methodology than ours and does not exploit the richness of our data obtained by mining images from customer reviews.

{\bf Visual Attributes}.
Research on attribute-based visual representations have received renewed attention by the computer vision community
in the past few years \cite{lampert2009learning, farhadi2009,parikh2011relative,vedaldi2014attributes}. Attributes are usually referred as semantic properties of objects or scenes that
are shared across categories. 
Among other applications, attributes have been used for zero-shot learning \cite{lampert2009learning},
image ranking and retrieval \cite{siddiquie2011image,kovashka2012whittlesearch,huang2014circle}, fine-grained categorization \cite{branson2010}, scene understanding \cite{scene2012},
and sentence generation from images \cite{baby2011}.

Related to our application domain, Kovashka et al \cite{kovashka2012whittlesearch} developed
a system called ``WhittleSearch'', which is able to answer queries such as ``Show me shoe images like these, but sportier''.
They used the concept of relative attributes proposed by Parikh and Grauman \cite{parikh2011relative} for relevance feedback.
Attributes for clothing have been explored in several recent papers \cite{chen2012describing,qiang2015,bourdev2011}. They allow users to search visual content
based on fine-grained descriptions, such as a ``blue striped polo-style shirt''.

Attribute-based representations have also shown compelling results for matching images of people across domains \cite{shi2015, nusreid2014}.
The work by Donahue and Grauman \cite{rationales2011} demonstrates that richer supervision conveying annotator rationales based on visual attributes, can be considered as a form of
privileged information \cite{vapnik2009new}. Along this direction, in our work, we show that cross-domain image retrieval
can benefit from feature learning that simultaneously optimizes a loss function that takes into account visual similarity and attribute classification.

{\bf Deep Learning}.
Deep convolutional neural networks have achieved dramatic accuracy improvements in many areas of computer vision \cite{krizhevsky2012imagenet, girshick2014rich,sun2015}.
The work of Zhang et al \cite{panda2014} combined poselet classifiers \cite{bourdev2011} with convolutional nets to achieve compelling results in human attribute prediction.
Sun et al \cite{sun2015} discovered that attributes can be implicitly encoded in high-level features of networks for identity discrimination. 
In our work, we instead explicitly use attribute prediction as a regularizer in deep networks for cross-domain image retrieval.

Existing approaches for image retrieval based on deep learning have outperformed previous methods based on other image representations \cite{neuraleccv2014}.
However, they are not designed to handle the problem of cross-domain image retrieval. 
Several domain adaptation methods based on deep learning have been recently proposed  \cite{oneshotJia2013,chopra2013dlid}. Related to our work, Chen et al \cite{qiang2015} uses a double-path network
with alignment cost layers for attribute prediction. In contrast, our work addresses the problem of cross-domain retrieval,
proposing a novel network architecture that learns effective features for measuring visual similarity across domains.

\begin{table}
\begin{center}
\scalebox{0.85}{
\begin{tabular}{|l|l|c|}
\hline
Attribute categories & Examples (total number) \\
\hline 
Clothes Button & Double Breasted, Pullover, ... (12) \\
Clothes Category & T-shirt, Skirt, Leather Coat ... (20) \\
Clothes Color & Black, White, Red, Blue ... (56) \\
Clothes Length & Regular, Long, Short ... (6) \\
Clothes Pattern & Pure, Stripe, Lattice, Dot ... (27) \\
Clothes Shape & Slim, Straight, Cloak, Loose ... (10) \\
Collar Shape & Round, Lapel, V-Neck ... (25) \\
Sleeve Length & Long, Three-quarter, Sleeveless ... (7) \\
Sleeve Shape & Puff, Raglan, Petal, Pile ... (16) \\
\hline
\end{tabular}
}
\end{center}
\caption{Clothing attribute categories and example values. The number in brackets is the total number of values for each category.}
\vspace{-0.1in}
\label{tab:clothes_attributes}
\end{table}

\begin{figure}
\begin{center}
\includegraphics[width=0.40\textwidth,keepaspectratio]{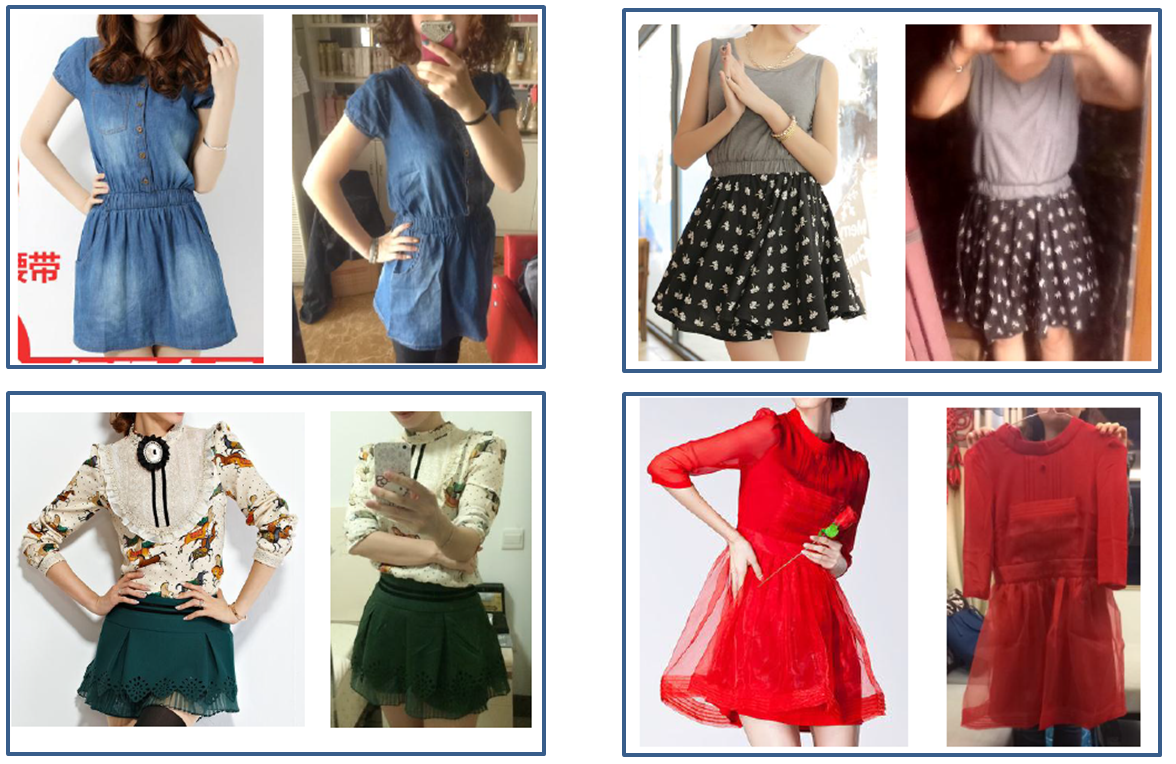}
\end{center}
\vspace{-0.1in}
\caption{Some examples of online-offline image pairs, containing images of different human pose,  illumination, and varying background. Particularly, the offline images contain many selfies with high occlusion.}
\label{fig:online_offline_pairs}
\vspace{-0.1in}
\end{figure}

\section{Data Collection}
\label{sec:data}
We have collected about 453,983 online upper-clothing images in high-resolution (about $800 \times 500$ on average) from several online-shopping websites. Generally, each image contains a single frontal-view person.
From the surrounding text of images,  semantic attributes (\eg, clothing color, collar shape, sleeve shape, clothing style) are extracted and parsed into \textless\textit{key}, \textit{value}\textgreater ~pairs, where each \textit{key} corresponds to an attribute category (\eg, color), and the \textit{value} is the attribute label (\eg, red, black, white, \etc). 
There are 9 categories of clothing attributes extracted from the websites and the total number of attribute values is 179. As an example, there are 56 values for the color attribute.

\begin{figure}
\begin{center}
\includegraphics[width=0.40\textwidth,keepaspectratio]{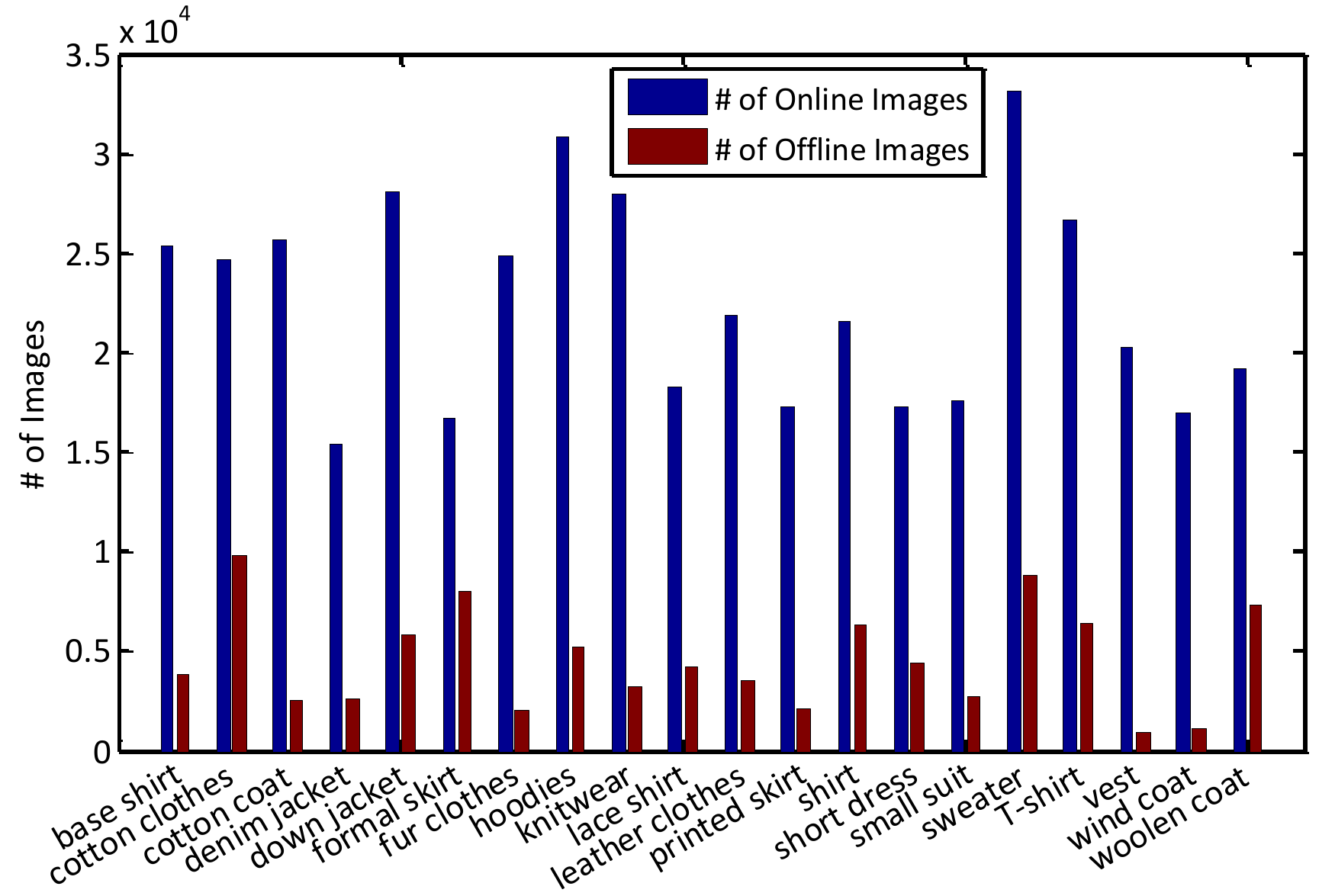}
\end{center}
\vspace{-0.2in}
\caption{The distribution of online-offline image pairs.}
\label{fig:clothes_distribution}
\vspace{-0.1in}
\end{figure}

The specified attribute categories and example attribute values are presented in Table~\ref{tab:clothes_attributes}.
This large-scale dataset annotated with fine-grained clothing attributes is used to learn a powerful semantic representation of clothing, as we will describe in the next section.

Recall that the goal of our retrieval problem is to find the online shopping images that correspond to a given query photo in the ``street'' domain uploaded by the user.
To analyze the discrepancy between the images in the shopping scenario (online images) and street scenario (offline images), we collect a large set of offline images with their online counterparts. The key insight to collect this dataset is that there are many customer review websites where users post photos of the clothing they have purchased. As the link to the corresponding clothing images from the shopping store is available, it is possible to collect a large set of online-offline image pairs. 

We initially crawled 381,975 online-offline image pairs of different categories from the customer review pages.
Then, after a data curation process, where several annotators helped removing unsuitable images, the data was reduced to  91,390 image pairs. For each of these pairs, fine-grained clothing attributes were extracted from the online image descriptions. Some examples of cropped online-offline image pairs are presented in Figure \ref{fig:online_offline_pairs}.
As can be seen, each pair of images depict the same clothing, but in different scenarios, exhibiting variations in pose, lighting, and background clutter. The distribution of the collected online-offline images is illustrated in Figure \ref{fig:clothes_distribution}.
Generally, the number of images of different categories in both scenarios are almost in the same order of magnitude, which is helpful for training the retrieval model.


In summary, our dataset is suitable to the clothing retrieval problem for several reasons.
First, the large amount of images enables effective training of retrieval models, especially deep neural network models.
Second, the information about fine-grained clothing attributes  allows learning of semantic representations of clothing.
Last but not least, the online-offline images pairs bridge the gap between the shopping scenario and the street scenario, providing rich information for real-world applications.
\vspace{-0.05in}

\section{Technical Approach}
\vspace{-0.05in}

\begin{figure*}
\begin{center}
\includegraphics[width=0.90\textwidth,keepaspectratio]{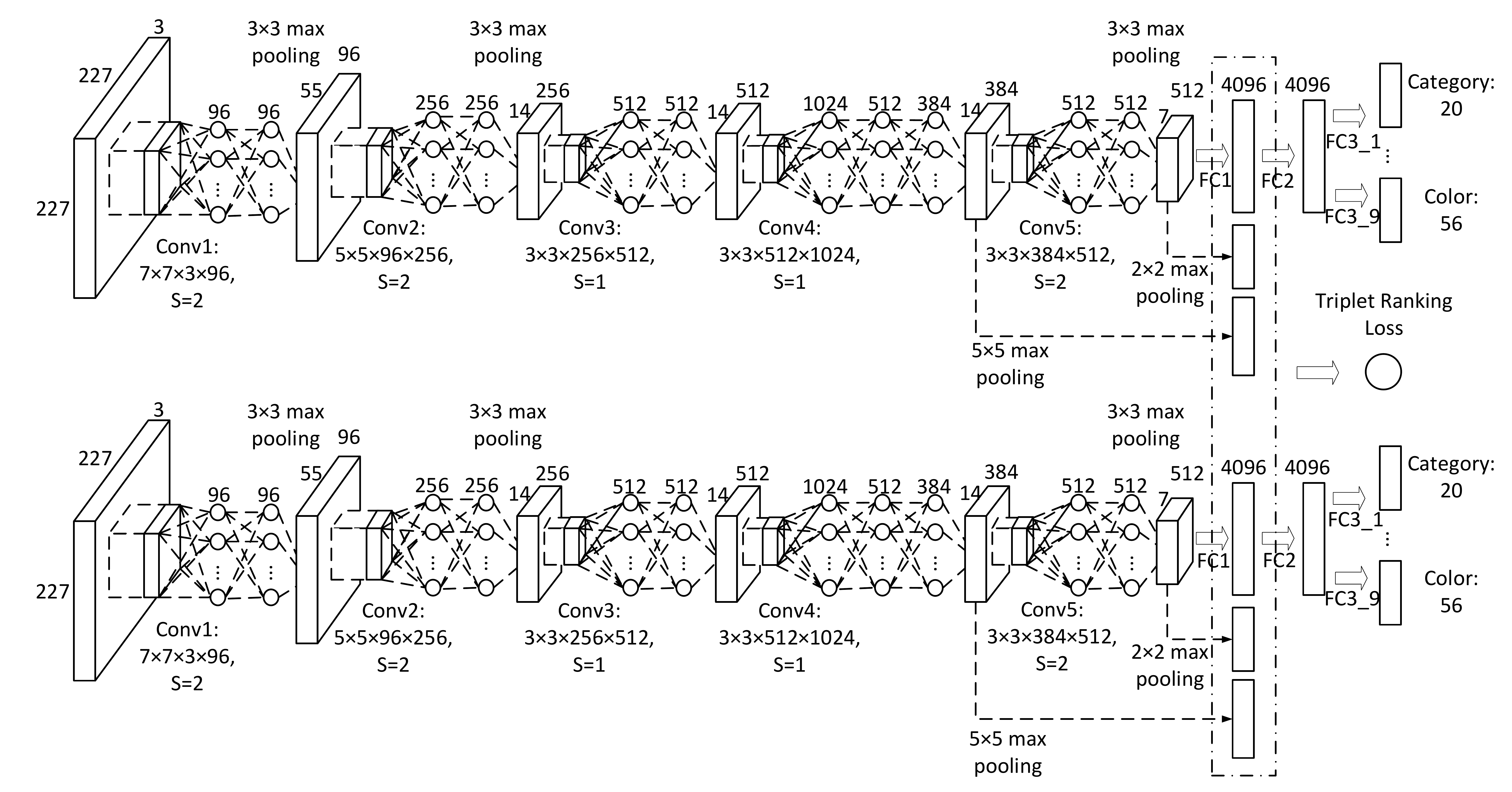}
\end{center}
\vspace{-0.2in}
\caption{The specific structure of DARN, which consists of two  sub-networks for images of the shopping scenario and street scenario, respectively. In each sub-network, it contains a NIN network, including all the convolutional layers, followed by two fully connected layers.   The tree-structure layers are put on top of each network for attribute learning. The output features of each sub-network, \ie, FC1, Conv4-5, are concatenated and fed into the triplet ranking loss across the two sub-networks.}
\label{fig:dan}
\vspace{-0.1in}
\end{figure*}

The unique dataset introduced in the previous section serves as the fuel to power up our
attribute-driven feature learning approach for cross-domain retrieval. Next we describe the main components of our proposed approach, and how they are assembled to create a real-world cross-domain clothing retrieval system.

\subsection{Dual Attribute-aware Ranking Network}
In this section, the Dual Attribute-aware Ranking Network (DARN) is introduced for retrieval feature learning. Compared to existing deep features \cite{jia2014caffe, donahue2013decaf}, DARN simultaneously integrates semantic attributes with visual similarity constraints into the feature learning stage, while at the same time modeling the discrepancy between domains.

\vspace{0.1in}
{\bf Network Structure}. 
\label{sec:NIN_structure}
The structure of DARN is illustrated in Figure \ref{fig:dan}.  Two sub-networks with similar Network-in-Network (NIN) models \cite{linmin2013nin} are constructed as its foundation. During training, the images from the online shopping domain are fed into one sub-network, and the images from the street domain are fed into the other. Each sub-network aims to represent the domain-specific information and generate high level comparable features as  output. 
The NIN model in each sub-network consists of five stacked convolutional layers followed by  MLPConv layers as defined in \cite{linmin2013nin}, and two fully connected layers (FC1, FC2).
To increase the representation capability of the intermediate layer, the fourth layer, named Conv4, is followed by two MLPConv layers. 

 On top of each sub-network, we add  tree-structured fully-connected layers to encode information about semantic attributes. Given the semantic features learned by the two sub-networks, we further impose a triplet-based ranking loss function, which separates the dissimilar images with a fixed margin under the framework of {\em learning to rank}.
The details of semantic information embedding and the ranking loss  are introduced next.


\vspace{0.1in}
{\bf Semantic Information Embedding}.
In the clothing domain, attributes often refer to the specific description of certain parts (\eg, collar shape, sleeve length) or clothing (\eg, clothes color, clothes style).
Complementary to the visual appearance, this information can be used to form a powerful semantic representation for the clothing retrieval problem. 
To represent the clothing in a semantic level, we design tree-structure layers to comprehensively capture the information of attributes and their full relations.

Specifically, we transmit the FC2 response of each sub-network to several branches, where each branch represents a fully-connected network to model each attribute separately. In this tree-structured network, the visual features from the low-level layers are shared among attributes; while the
semantic features from the high-level layers are learned separately.
The neuron number in the output-layer of each branch equals to the number of corresponding attribute values (see Table \ref{tab:clothes_attributes}).
Since each attribute has a single value, the cross-entropy loss is used in each branch.
Note that the values of some attributes may be missing for some clothing images. In this case,  the gradients from the corresponding branches are simply set to zero.

During the training stage, the low-level representation of clothing images is extracted layer by layer.
As the activation transfers to the higher layers, the representation becomes more and more abstract.
Finally, the distinctive characteristic of each attribute is modeled in each branch.
In the back-propagation, the gradient of loss from each attribute w.r.t. the activation of FC2 layer are summed up and transferred back for weight update.
\vspace{0.1in}
{\bf Learning to Rank with Semantic Representation}. In addition to encoding the semantic representation, we apply the learning to rank framework on DARN for  retrieval feature learning.
Specifically, the triplet-based ranking loss is used to constrain the feature similarity of image triplets.
Denoting $a$ and $b$ the features of an offline image and its corresponding online image respectively, the objective function of the triplet ranking loss is:
\begin{equation}
\vspace{-0.1in}
Loss(a, b, c) = max(0, m + dist(a, b) - dist(a, c)),
\end{equation}
where $c$ is the feature  of the dissimilar online image, $dist(\cdot, \cdot)$ represents the feature distance, \eg, Euclidean distance, and $m$ is the margin, which is empirically set as 0.3 according to the average feature distance of image pairs.
Basically, this loss function imposes that the feature distance between an online-offline clothing pair should be less than that of the offline image and any other dissimilar online image by at least margin $m$.

In this way, we claim that the triplet ranking loss has two advantages.
First and obviously, the desirable ranking ordering can be learned by this loss function.
Second, as the features of online and offline images come from two different sub-networks, this loss function can be considered as the constraint to guarantee the comparability of features extracted from those two sub-networks, therefore bridging the gap between the two domains.

Similar to \cite{donahue2013decaf}, we found that the response of FC1 layer, \ie, the first fully connected layer, achieves the best retrieval result.
Therefore, the triplet ranking loss is connected to the FC1 layer for feature learning.
However, the response from the FC1 layer  encodes global features, implying that subtle local information may be lost, which is specially relevant for discriminating clothing images.
To handle this problem, we claim that local features captured by convolutions  should also be considered. Specifically, the max-pooling layer is used to down-sample the response of the convolutional layers into $3 \times 3 \times f_n$, where $f_n$ is the number of filters in the $n$-th convolutional layer.
Then, the down-sampled response is vectorized and concatenated with the global features.
Lastly, the triplet ranking loss is applied on the concatenated features of every triplet.
In our implementation, we select the pooled response map of Conv4 and Conv5, \ie, the last two convolutional layers, as local features.

\subsection{Clothing Detection}

As a pre-processing step, the clothing detection component aims to eliminate the impact of cluttered backgrounds by cropping the foreground clothing from images, before feeding them into DARN. Our method is an enhanced version of the R-CNN approach \cite{girshick2014rich}, which has recently achieved state-of-the-art results in object detection.

Analogous to the R-CNN framework, clothing proposals are generated by  selective search \cite{uijlings2013selective}, with some unsuitable candidates discarded by constraining the range of size and aspect ratio of the bounding boxes.
Similar to Chen  et al \cite{qiang2015}, we process the region proposals by a NIN model.  However, our model differs in the sense that we use the attribute-aware network with  tree-structured layers as described in the previous section, in order to embed semantic information as extra knowledge. We show in our experiments that this model yields superior results.

Based on the attribute-aware deep features,  support vector regression (SVR) is used to predict the intersection over union (IoU) of clothing proposals.
In addition,  strategies such as the discretization of IoU on training patches, data augmentation, and hard example mining, are used in our training process.
As post-processing, bounding box regression is employed to refine the selected proposals with the same features used for detection.


\subsection{Cross-domain Clothing Retrieval}
\label{sec:retrieval}

We now describe the implementation details of our complete system for cross-domain clothing retrieval.


\vspace{0.1in}
\textbf{Training Stage}.
The training data is comprised of online-offline clothing image pairs with fine-grained clothing attributes, as described in Section~\ref{sec:data}. The clothing area is extracted from all images using our clothing detector, and then the cropped images are arranged into triplets.



In each triplet, the first two images are the online-offline pairs, with the third image randomly sampled from the online training pool.
As the same clothing images have an unique ID, we sample the third online image until getting a different ID than the online-offline image pair.
Several such triplets construct a training batch, and the images in each batch are sequentially fed into their corresponding sub-network according to their scenarios. 
We then calculate  the gradients for each loss function (cross-entropy loss and triplet ranking loss) w.r.t. each sample.  The gradients from the triplet loss function are back propagated to each individual sub-network according to the sample domain.

We pre-trained our network as well as the baseline networks used in the experiments on the ImageNet dataset (ILSVRC-2014), as this yields improved retrieval results when compared to random initialization of parameters.



\vspace{0.1in}
\textbf{End-to-end Clothing Retrieval}. We have set up an end-to-end real-time clothing retrieval demo on our local server (will be published soon).
We also provide example retrieval results in the supplemental files. 
In our retrieval system, 200,000 online clothing images cropped by the clothing detector are used to construct our retrieval gallery. Given the cropped online images, the concatenated responses from FC1 layer, pooled Conv4 layer, and pooled Conv5 layer of one sub-network of DARN corresponding to shop scenario are used as the representation features. The same processes are operated on the query image, except that the other sub-network of DARN is used for retrieval feature extraction.
We then  $l_2$ normalize  the features from different layers for each image. The PCA is used to reduce the dimension of normalized features (17,920-D for DARN with Conv4-5) into 4,096-D, which conducts a fair comparison with other deep features using FC1 layer output only.
Based on the pre-processed features, the Euclidean distance between query and gallery images is used to rank the images according to the relevance to the query.

\section{Experiments}


\subsection{Experimental Setting}

\textbf{Dataset}:
For training the clothing detector,  7,700 online-offline images are sampled from our dataset as positives and labelled with bounding boxes.
The person-excluded images from the PASCAL VOC 2012 \cite{voc2012} detection task are used as negatives.
Another 766 images are annotated to test the detectors.

For the retrieval experiment, about 230,000 online images and 65,000 offline images are sampled for network training. In the training process, each offline image and its online counterpart are collected, with the dissimilar online image randomly sampled from the 230,000 online pool to construct a triplet.
Note that the third images in different epochs are shuffled to be different for the same online-offline pair. For testing, we used 1,717 online-offline image pairs.
To make the retrieval result convincing, the rest 200,000 online images are used as the retrieval gallery.

\textbf{Baselines}:
For clothing detection, we compare the performance of Deformable Part-based Model (\textbf{DPM}) \cite{felzenszwalb2010cascade} and different R-CNN versions with different models, 
 including AlexNet (\textbf{Pre-trained CNN}) \cite{krizhevsky2012imagenet}, \textbf{Pre-trained NIN}, and the Attribute-aware Network (\textbf{AN}).
To evaluate the contribution of SVR, we compare the performance of SVR and SVM based on the AlexNet.

For clothing retrieval, the approach using Dense-SIFT (DSIFT) + fisher vector (FV) is selected as traditional baseline.
Specifically, the bin size and stride for DSIFT are 8 and 4, respectively.
The descriptor dimension is reduced to 64 by PCA.
In the encoding step, two dictionaries with 64 and 128 centers are constructed, which lead to the 8,192 and 16,384 dimensions of FV representation.

To analyze the retrieval performance of deep features, we compare pre-trained networks including  AlexNet (\textbf{pre-trained CNN}) and \textbf{pre-trained NIN}.
We evaluate each individual component of our proposed approach.
We denote our overall solution as Dual Attribute-aware Ranking Network (\textbf{DARN}), the solution without dual structure as Attribute-aware Ranking Network (\textbf{ARN}), the solution without dual structure and the ranking loss function as Attribute-aware Network (\textbf{AN}). 

We further evaluate  the effectiveness of DARN in terms of different configurations w.r.t. the features used, \ie, \textbf{DARN} using the features obtained from FC1, \textbf{DARN with Conv4} using the features from FC1+Conv4, and \textbf{DARN with Conv4-5} using the features from FC1+Conv4+Conv5. It is worth noting that the dimension of all features are reduced to 4096 by PCA to have a fair comparison.

\textbf{Evaluation Metrics:} We used two metrics to measure the  performance of retrieval models. (1) the top-k retrieval accuracy in which we denote a hit if we find the exact same clothing in the top $k$ results otherwise a miss, and (2) Normalized Discounted Cumulative Gain ($NDCG@k$), considering $NDCG@k=\frac{1}{Z} \sum_{j=1}^{k} \frac{2^{rel(j)}-1}{log(j+1)}$, where $rel(j)$ is the relevance score of the $j^{th}$ ranked image, and $Z$ is a normalization constant.
The relevance score $rel(j)$ of query image and $j^{th}$ ranked image is the number of their matched attributes divided by the total number of query attributes.



\begin{table}
\begin{center}
\scalebox{0.9}{
\begin{tabular}{|l|c|c|c|}
\hline
Detection Model & Online AP & Offline AP & Top-20 Acc\\
\hline 
DPM & 0.049 & 0.017 &0.297\\
Pre-t CNN+SVM & 0.520 & 0.412 &0.560\\
Pre-t CNN+SVR & 0.545 & 0.452 &0.567\\
Pre-t NIN+SVR & 0.601 & 0.477 &0.588\\
AN+SVR &\textbf{0.744}  & \textbf{0.683} & \textbf{0.635}\\
\hline
\end{tabular}
}
\end{center}
\vspace{-0.1in}
\caption{AP of detection models on online-offline images and its corresponding top-20 retrieval accuracy on a subset of the data.}
\vspace{-0.1in}

\label{tab:clothes_detection}
\end{table}


 \subsection{Clothing Detection Improving Clothing Retrieval Performance}
We used Average Precision (AP) to evaluate clothing detection.
Since the detection performance is important to our network learning, a more strict IoU threshold, \ie, 0.7, is selected.
The AP of detection results on online and offline images is presented in Table \ref{tab:clothes_detection}.
Generally, the performance of every detector on the online images is better than that on offline images, which indicates the complexity of offline images.
We can observe that our proposed AN with SVR is superior than other baselines.
DPM achieves the lowest AP, which may be due to less discriminative features and its incapability to handle clothing with huge distortion.
By comparing the performance of CNN with SVM and SVR, we can find the effectiveness of SVR in the R-CNN framework.
Furthermore, the detection performance is further improved by replacing the CNN with pre-trained NIN.
Lastly, the AN with SVR achieves 74.4\% and 68.3\% AP on the online and offline images respectively, which is significantly better than the runner-up.

To evaluate the impact of various detectors on retrieval, we compare the top-20 retrieval accuracy of DARN with Conv4-5 by feeding different detection results.
We sampled 10,000 online images from the full set as retrieval gallery for this test.
The results are presented in Table \ref{tab:clothes_detection}.
As can be seen, more precise detection leads to more accurate retrieval results.
\begin{figure}
\includegraphics[width=0.45\textwidth,keepaspectratio]{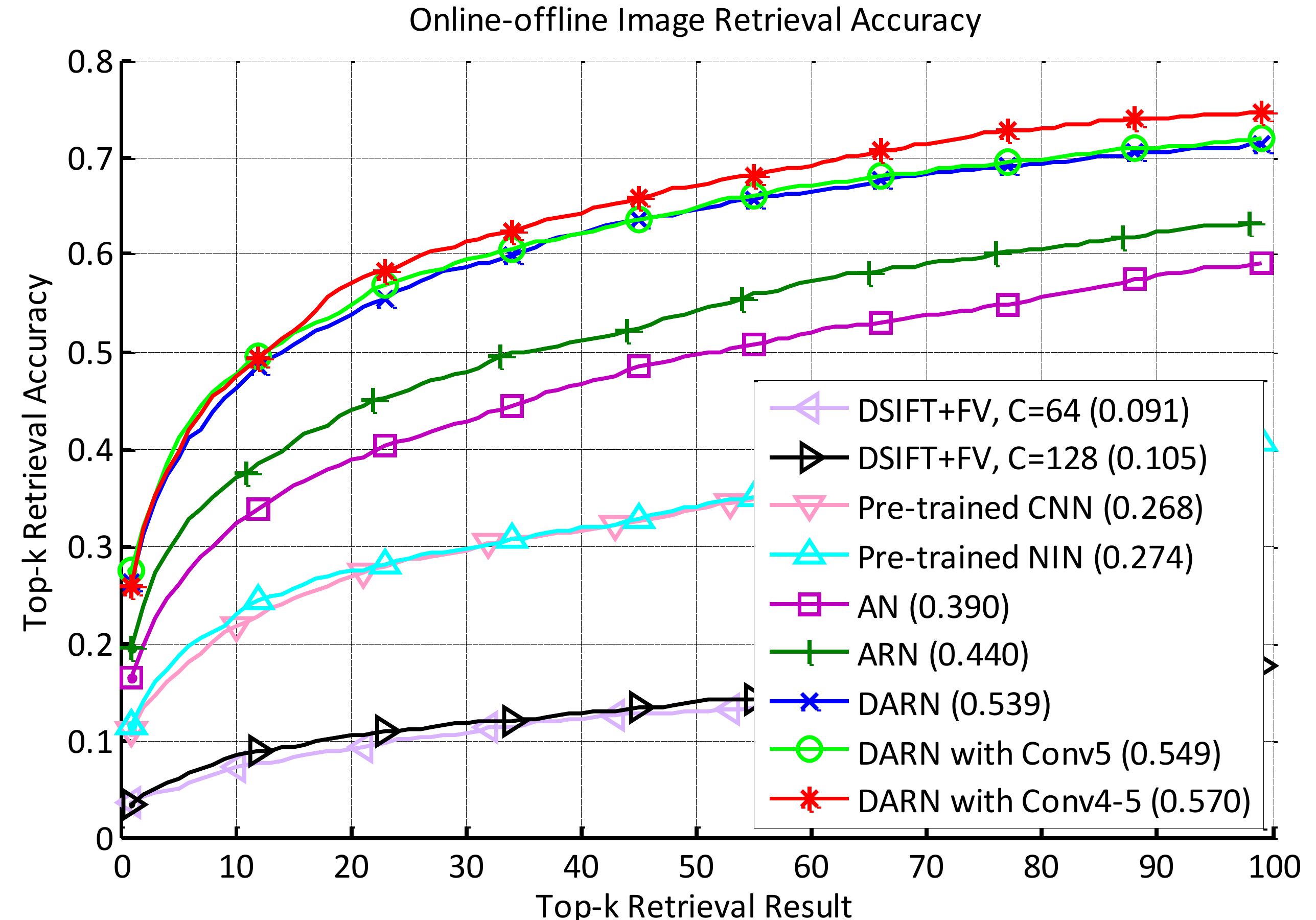}
\caption{The top-k retrieval accuracy on 200,000 retrieval gallery. The number in the parentheses is the top-20 retrieval accuracy.}
\label{fig:top_k_accuracy}
\end{figure}

\begin{figure}
\begin{center}
\includegraphics[width=0.45\textwidth,keepaspectratio]{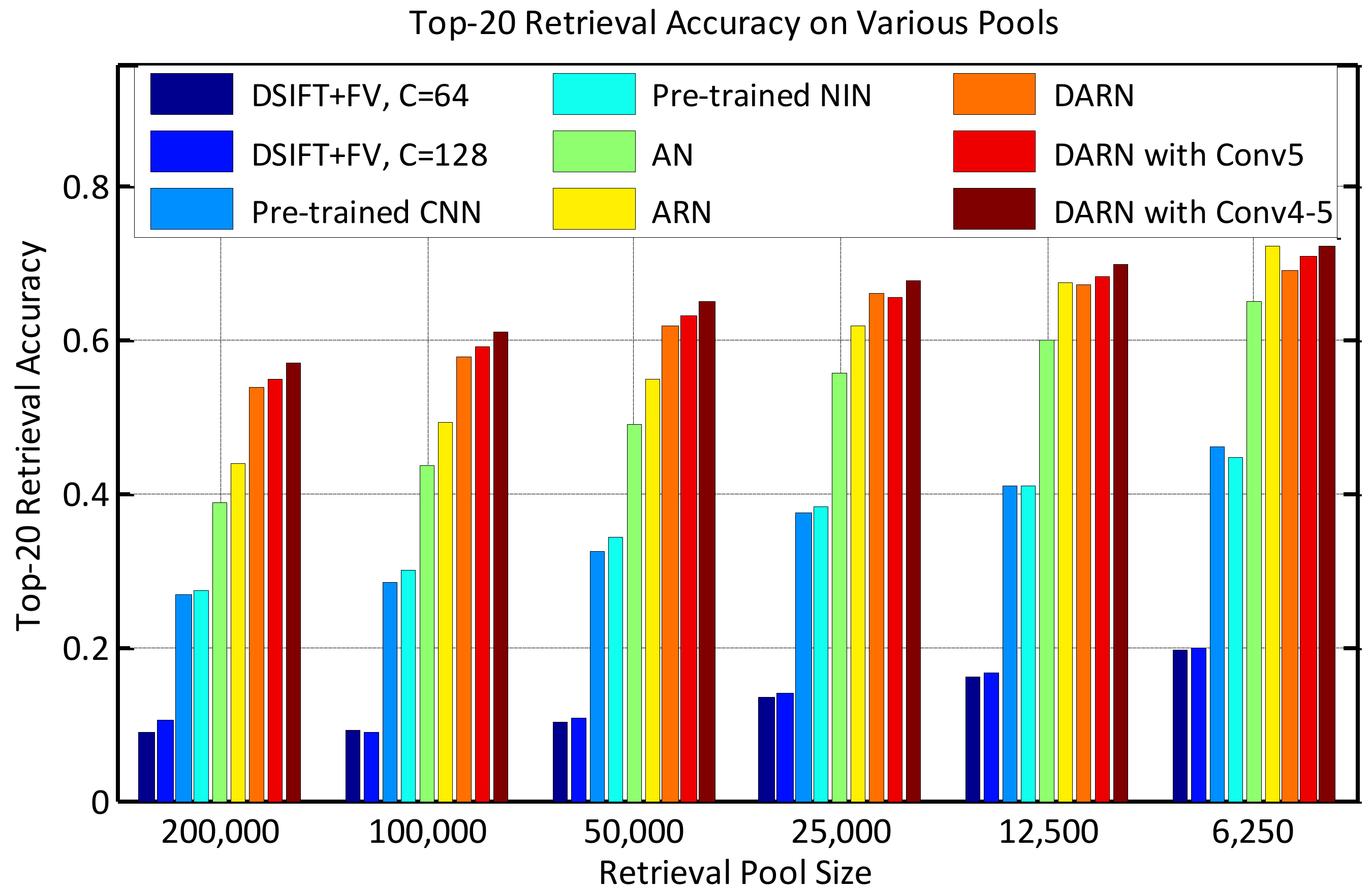}
\end{center}
\vspace{-0.1in}
\caption{The top-20 retrieval accuracy on different sizes of retrieval galleries.}
\label{fig:top_k_var_pool}
\vspace{-0.1in}
\end{figure}

\begin{figure*}
\begin{center}
\includegraphics[width=0.90\textwidth,keepaspectratio]{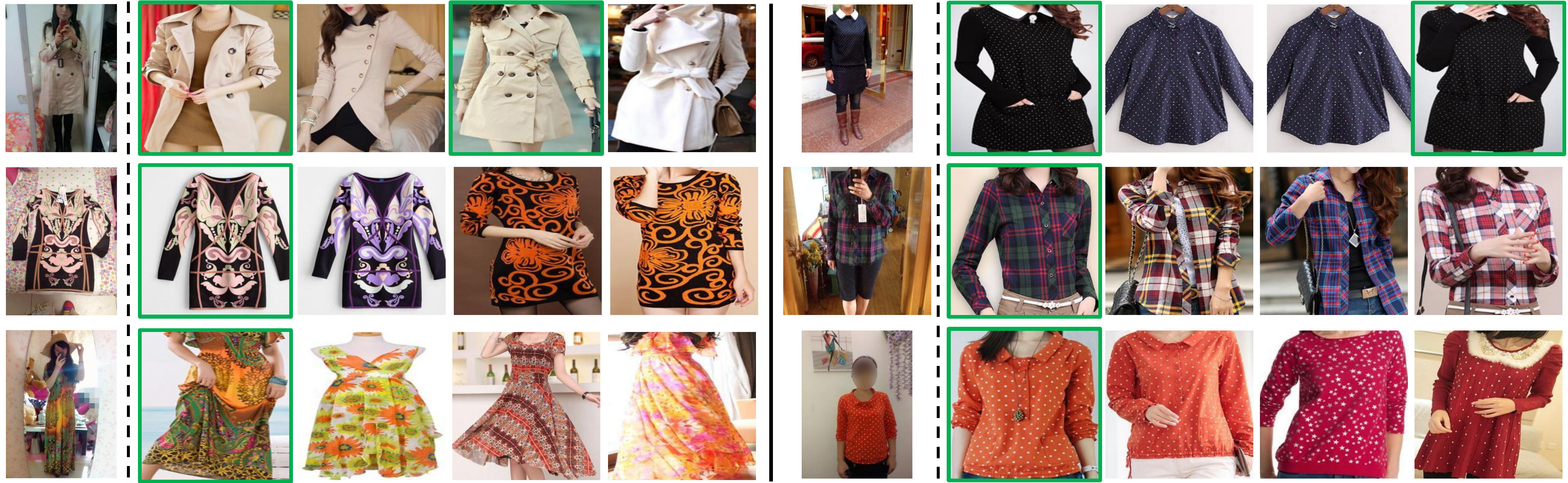}
\end{center}
\vspace{-0.1in}
\caption{The top-4 retrieval result of DARN with Conv4-5. The images in first column are the queries, and the retrieved images with green boundary are the same clothing images. Best viewed in original pdf file.}
\label{fig:retrieval_result_vis}
\vspace{-0.2in}
\end{figure*}

\subsection{Cross-domain Clothing Retrieval Evaluation}
We give full detailed top-k retrieval accuracy results for different baselines as well as our proposed methods  in Figure \ref{fig:top_k_accuracy}. We vary $k$ as the tuning parameter as it is an important indicator for a real system.  We also list  the top-20 retrieval accuracy of each model in the parentheses.

Compared to the baselines, we notice that  all the deep features significantly outperform the traditional features, \ie, Dense-SIFT with FV encoding. For the deep features, the top-k accuracy of pre-trained NIN is slightly better than that of pre-trained CNN. Based on the pre-trained NIN, we evaluate the contributions of tree-structured layers, triplet ranking, and dual-structure.

Generally, the retrieval performance is gradually improved by applying the NIN structure, semantic information, learning to rank framework, and the dual-structure. The top-20 retrieval accuracy of AN increases 11.6\% after fine-tuning on pre-trained NIN with attributes.  This attests the effectiveness of attributes for image retrieval.
By introducing the triplet ranking loss, the top-20 accuracy of ARN achieves another 5.0\% increment.

Compared with a single model, the dual-structure network greatly improves the retrieval performance, i.e., the top-20 retrieval accuracy of DARN improves 9.9\% when compared with ARN.
The retrieval performance also slightly benefits from the local features, which can be observed by comparing the DARN and DARN with local features, \ie, DARN with Conv5 and DARN with Conv4-5.
Some retrieval examples by DARN with Conv4-5 are illustrated in Figure \ref{fig:retrieval_result_vis}. 

\subsection{Attribute-aware Clothing Retrieval Evaluation}
One key advantage of the proposed approach is the attribute-aware nature. The learned features have strong semantic meaning. Therefore, we should expect that the retrieval result should present strong attribute-level matching in terms of retrieval accuracy.  

To evaluate this argument, we use NDCG@K to calculate the attribute-level matching. More specifically, we define the relevance score in NDCG as the attribute matching between the query and retrieval results divided by the total number of query attributes.  
We present the result in Table \ref{tab:attr_ndcg}. Compared with traditional features, the retrieval result of deep features contains more similar attributes to the queries. 

\subsection{Showing the Robustness: Performance vs. Retrieval Gallery Size}
To further demonstrate the robustness our method, we show the top-20 retrieval accuracy of different retrieval models by tuning the retrieval gallery size in Figure \ref{fig:top_k_var_pool}.

We calculate the accuracy increment ratio of some representatives to evaluate the robustness of features.
Intuitively, the smaller increase ratio indicates the better robustness of features.
Specifically, the top-20 retrieval accuracy of traditional features, pre-trained NIN, ARN, and DARN increase by 115.4\%, 63.8\%, 64.5\%, and 28.2\% from largest retrieval gallery to smallest gallery, respectively.
This observation demonstrates that the DARN can learn much more robust features than the baselines.

\begin{table}
\begin{center}
\scalebox{0.9}{
\begin{tabular}{|l|c|}
\hline
Retrieval Model & NDCG@20 \\
\hline 
 DSIFT + FV, C = 64 & 0.290 \\
 DSIFT + FV, C = 128 & 0.289 \\
 Pre-trained CNN & 0.367 \\
 Pre-trained NIN & 0.370 \\
 AN & 0.415 \\
 ARN & 0.442 \\
 DARN & 0.494 \\
 DARN with Conv5 & 0.499 \\
 DARN with Conv4-5 & 0.505 \\
\hline
\end{tabular}
}
\end{center}
\vspace{-0.1in}
\caption{The NDCG@20 result evaluating the attribute level match on 200,000 retrieval gallery.}
\vspace{-0.1in}
\label{tab:attr_ndcg}
\end{table}

\subsection{System Running Time}

Our retrieval system runs on a server with the Intel i7-4930K CPU (@ 3.40GHz) with 12 cores and 65 GB RAM memory, with two GTX Titan GPU cards.
On average, the attribute-aware ranking feature extraction process costs about 13 seconds per 1,000 images.
Given a cropped query, it costs about 0.21 second for feature extraction and clothing retrieval in our retrieval experiment.


\section{Conclusions}
We have presented the Dual Attribute-aware Ranking Network for the problem of cross-domain image retrieval. Different from previous approaches, our method simultaneously embeds semantic attribute information and visual similarity constraints into the feature learning stage, while modeling the discrepancy of the two domains.  We demonstrate our approach in a practical real-world clothing retrieval application, showing substantial improvement over other baselines. In addition, we created a unique large-scale clothing dataset which should be useful to many other applications.



{\footnotesize	
\bibliographystyle{ieee}
\bibliography{egbib}
}

\begin{appendices}



\section{Clothing Detection}

\begin{figure}
\begin{minipage}{.45\textwidth}
  \begin{center}
  \includegraphics[width=\textwidth,keepaspectratio]{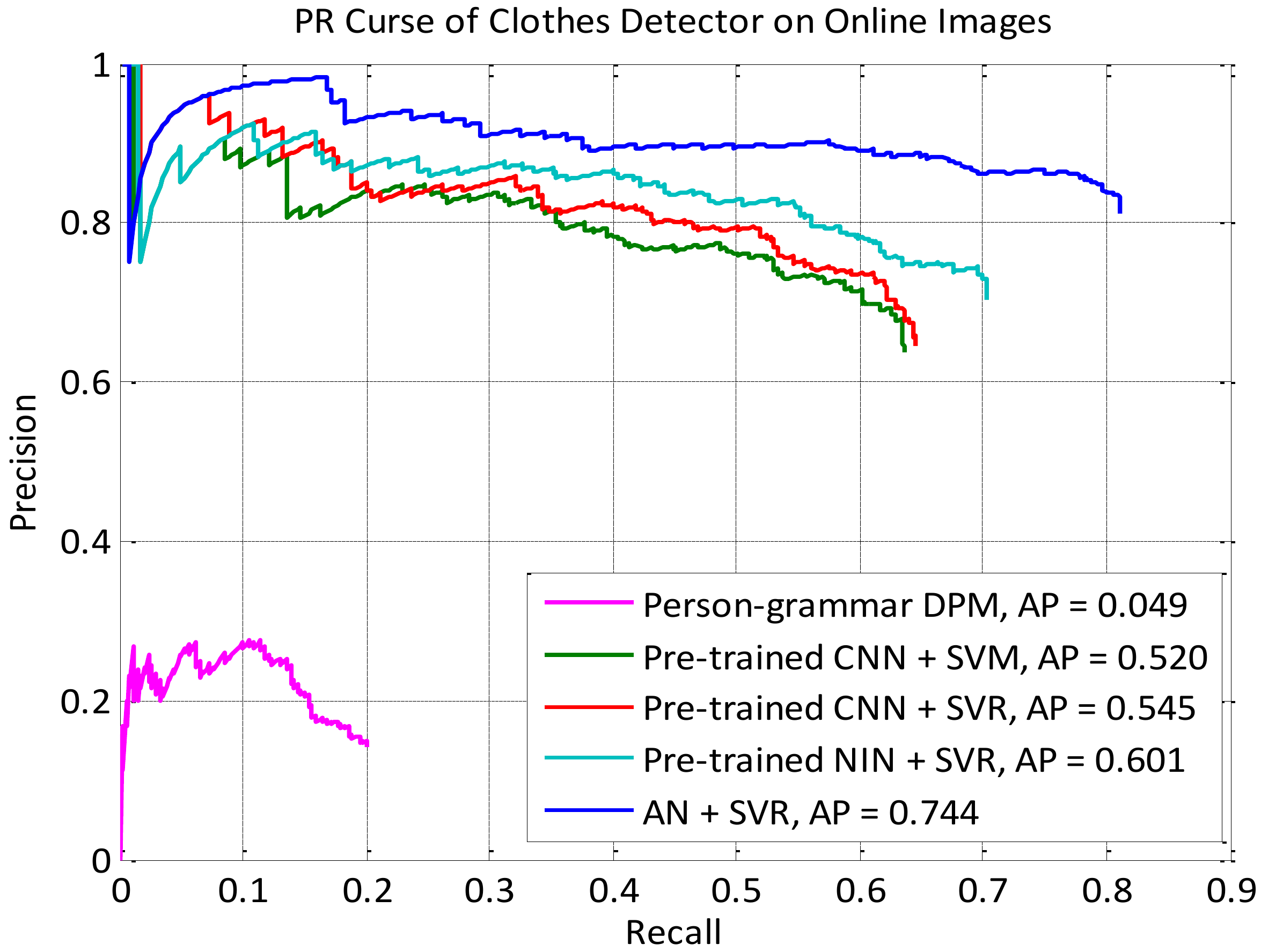}
  \end{center}
\end{minipage}
\begin{minipage}{.45\textwidth}
  \begin{center}
  \includegraphics[width=\textwidth,keepaspectratio]{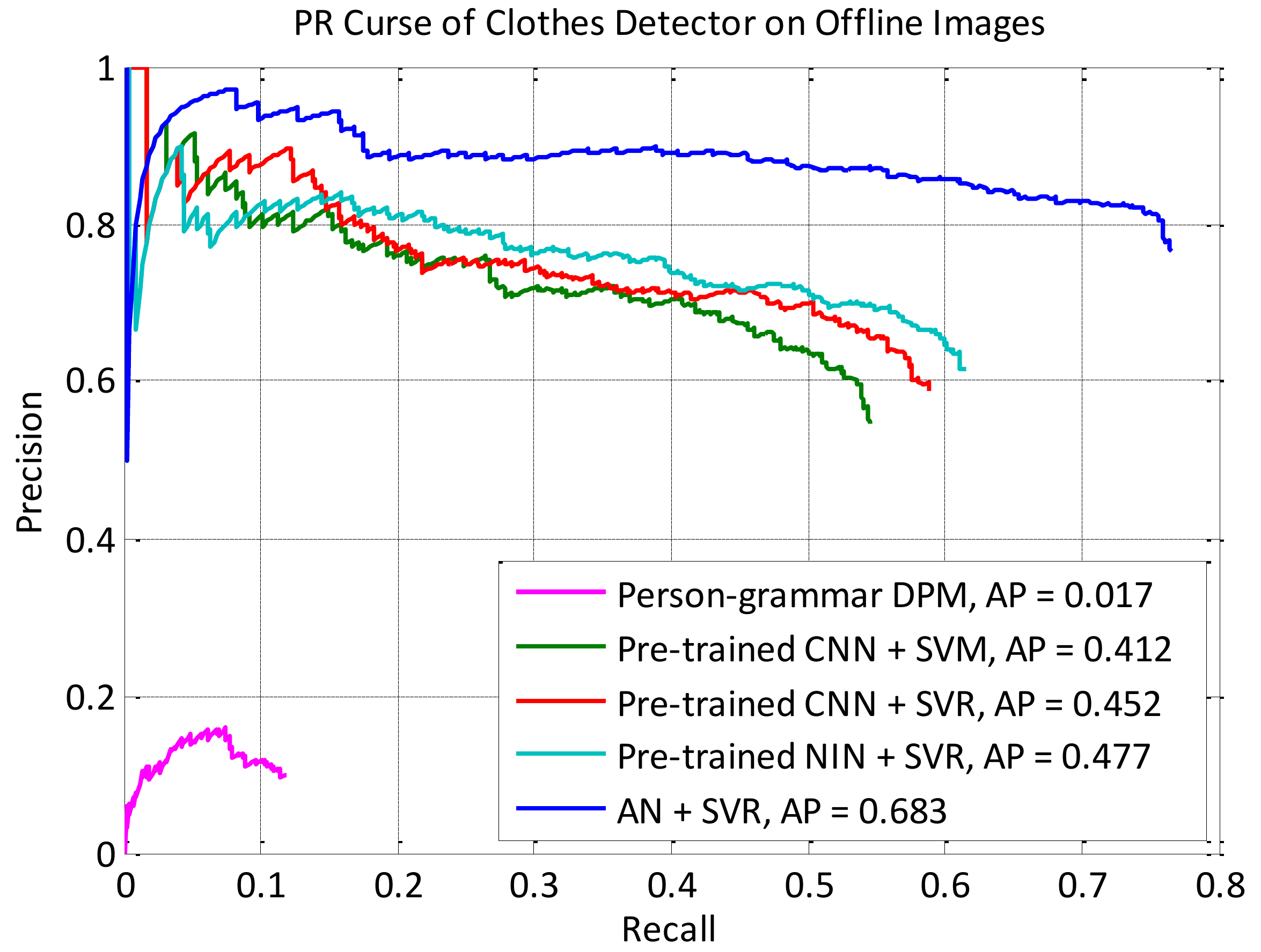}
  \end{center}
\end{minipage}
\caption{The Precision-recall curves of upper-clothes detection result on online images (top) and offline images (bottom).}
\label{fig:clothes_detection}
\vspace{-0.2in}
\end{figure}

The precision-recall curves of various detectors on the online-offline images are presented in Figure \ref{fig:clothes_detection}.
Generally, the performance of R-CNN framework outperforms the traditional framework, \ie, DPM.
Specifically, the performance of R-CNN framework is gradually enhanced by applying the support vector regression (SVR), NIN model, and the attribute-aware fine-tuning strategy.
Particularly, the R-CNN with AN + SVR achieving 74.4\% and 68.3\% average precision (AP) on online and offline images respectively is much superior than the other baselines.

\begin{figure*}
\begin{center}
\includegraphics[width=0.9\textwidth,keepaspectratio]{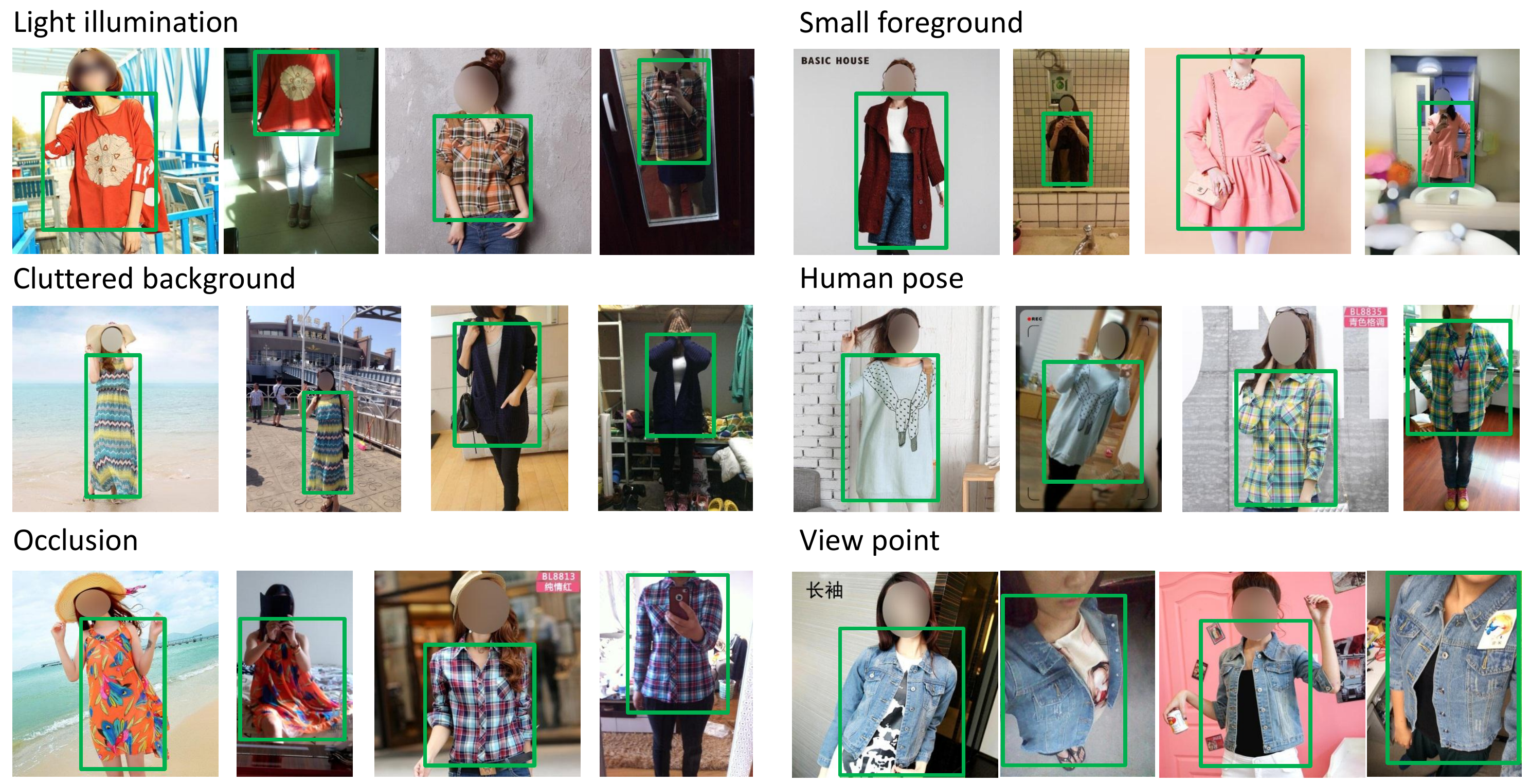}
\end{center}
\caption{Some detection results of R-CNN framework with AN + SVR.}
\label{fig:detection_vis}
\end{figure*}

We present some detection results of R-CNN with AN + SVR in Figure \ref{fig:detection_vis}.
As can be seen, the R-CNN framework with AN + SVR is robust in detecting online and offline images with different illuminations, cluttered background, different human poses, huge occlusion, and different viewpoints.

\section{Clothing Retrieval}

\begin{figure*}
\begin{center}
\includegraphics[width=0.9\textwidth,keepaspectratio]{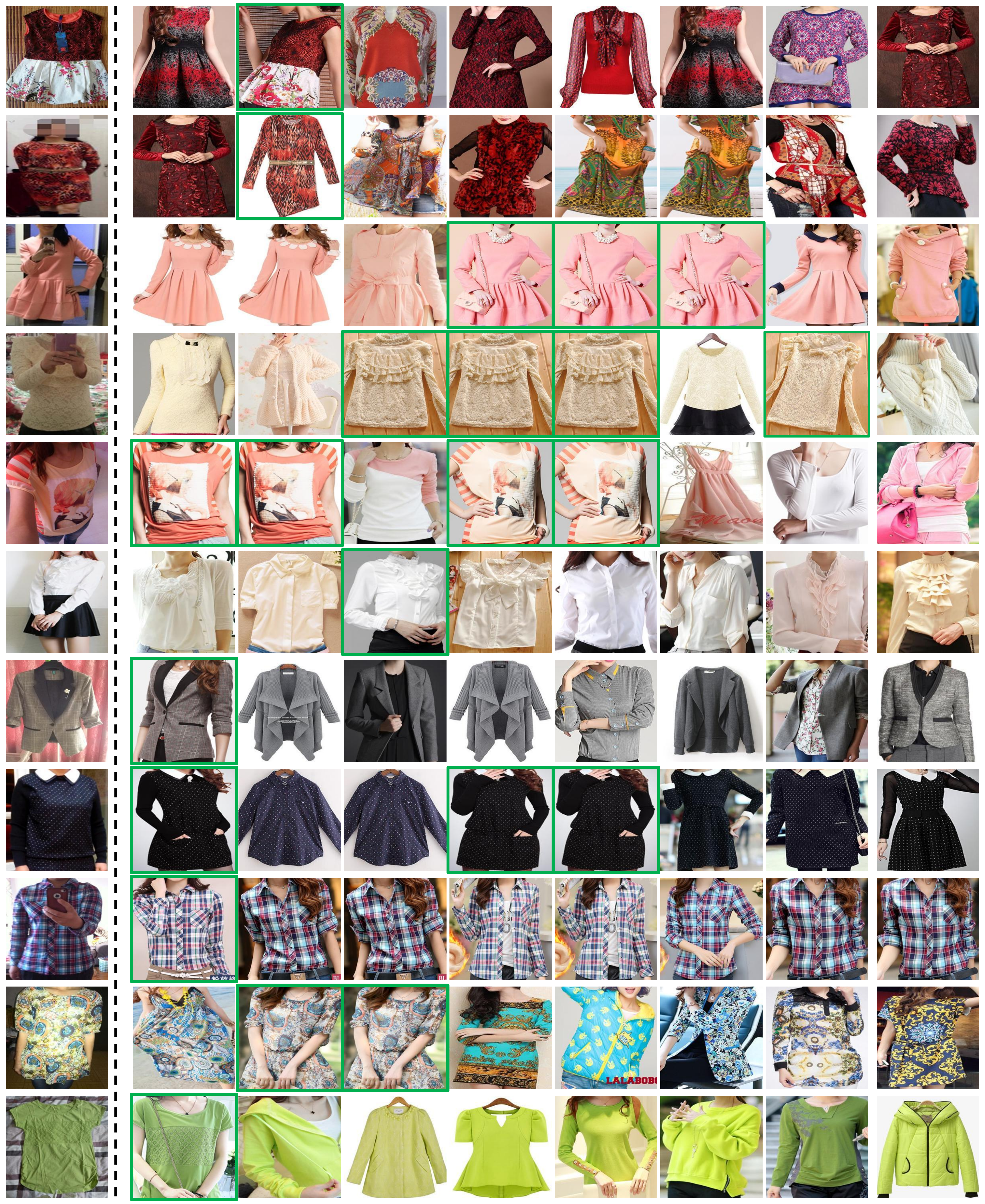}
\end{center}
\caption{Top-8 retrieval results of our proposed DARN with Conv4-5 model. The first column are the query images.}
\label{fig:retrieval_vis}
\end{figure*}

In our retrieval system, the foreground clothing is cropped by the R-CNN framework with AN + SVR before retrieval feature extraction.
Some retrieval results of DARN with Conv4-5 are illustrated in Figure \ref{fig:retrieval_vis}, from which we can observe that given a query image in street scenario, our proposed model can not only retrieve the same clothing in shop scenario, but also rank the online images with similar attributes as top results.

\end{appendices}

\end{document}